# Visual-Language Model Knowledge Distillation Method for Image Quality Assessment

Yongkang Hou, Jiarun Song

School of Telecommunications Engineering, Xidian University, Xi'an, China

*Abstract*—Image Quality Assessment (IQA) is a core task in computer vision. Multimodal methods based on vision-language models, such as CLIP, have demonstrated exceptional generalization capabilities in IQA tasks. To address the issues of excessive parameter burden and insufficient ability to identify local distorted features in CLIP for IQA, this study proposes a visual-language model knowledge distillation method aimed at guiding the training of models with architectural advantages using CLIP's IQA knowledge. First, quality-graded prompt templates were designed to guide CLIP to output quality scores. Then, CLIP is fine-tuned to enhance its capabilities in IQA tasks. Finally, a modality-adaptive knowledge distillation strategy is proposed to achieve guidance from the CLIP teacher model to the student model. Our experiments were conducted on multiple IQA datasets, and the results show that the proposed method significantly reduces model complexity while outperforming existing IQA methods, demonstrating strong potential for practical deployment.

## I. Introduction

In recent years, research on vision-language pre-trained models has made significant breakthroughs, particularly with OpenAI's CLIP [1]. By training on 400 million image-text pairs scraped from the internet using contrastive learning, its image encoder not only captures global semantic information from images but also gains cross-modal understanding capabilities, greatly advancing multi-modal learning and image understanding technologies. Work applying CLIP to IQA tasks has also emerged. CLIP-IQA [2] was the first to use CLIP for assessing image quality and abstract perception without requiring task-specific training, based on two antonymous prompt strategies learned through subjective quality scores; LIQE [3] introduced a multi-task learning approach that utilizes textual prompts incorporating distortion and scene information. By computing joint probabilities through cosine similarity, the method effectively infers predictions for each task.

Since CLIP has learned rich image-related knowledge through large-scale pre-training, it possesses outstanding capabilities in image semantic understanding, enabling it to demonstrate strong generalization capabilities across various visual tasks, including IQA. However, CLIP still faces numerous challenges in practical applications. First, CLIP has an excessive number of parameters, making it unsuitable for deployment and application on resource-constrained edge computing devices or mobile terminals. Second, the image encoder architecture adopted by CLIP is typically based on Vision Transformer [4], which focuses on global feature modeling and excels at capturing macro-level semantic structures and overall semantic information in images [5]. However, IQA not only needs to focus on the overall semantic structure of images but also requires accurate identification of local distortions and detail defects within images. Therefore, lightweight model architectures with integrated local feature modeling are preferable for IQA tasks.

How to leverage the advantages of the model architecture while effectively utilizing CLIP's rich pre-trained knowledge remains a challenge. On one hand, the dataset used for training is closed-source, making it difficult for ordinary researchers to obtain; on the other hand, CLIP adopts a multimodal contrastive learning paradigm, with all its datasets being image-text matching data, which fundamentally differs from the single-modal supervised training of traditional image encoders in terms of data structure and task objectives.

Knowledge distillation [6], as an effective method for knowledge transfer and model compression, offers a potential solution to this problem. We propose a visual-language model knowledge distillation method. First, a prompt-based design method is introduced to guide CLIP in completing the IQA task, enabling it to output a predicted quality score for the image. Then, a robust fine-tuning strategy tailored for CLIP is applied, where only its image encoder is fine-tuned on the IQA dataset to enhance its ability to perceive image quality-related features. Finally, through our proposed modality-adaptive knowledge distillation strategy, we first design a student model aligned with the CLIP modality. The enhanced CLIP serves as the teacher model, and a cosine-annealing-based weight scheduling strategy is adopted to balance soft and hard labels, effectively guiding the student with the teacher's IQA knowledge.

The main contributions of this study include:

- Addressing the challenges of inaccessible CLIP pre-training data and mismatched multi-modal and single-modal training paradigms, we propose a modality-adaptive knowledge distillation strategy to enable the CLIP teacher model's IQA knowledge to guide the student model.

- Through experiments on various architectures and student models with different parameter sets, it was demonstrated that the proposed method significantly improves the performance of the student model in IQA tasks compared to single-modal supervised training, while significantly reducing model complexity, demonstrating strong practical application potential.

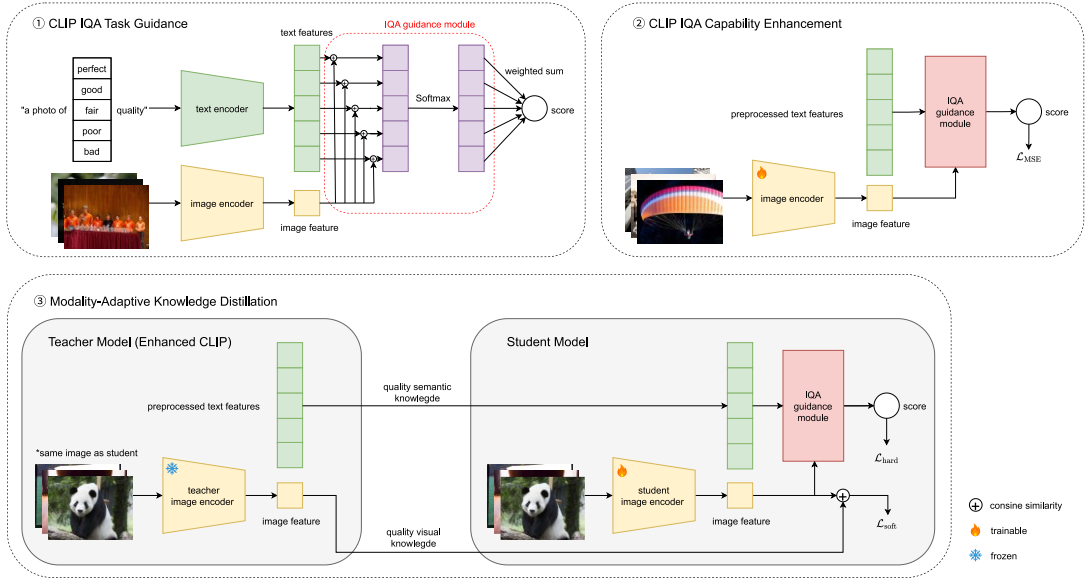

Fig. 1. Overall Architecture

## II. METHODOLOGY

A visual-language model knowledge distillation method is proposed for IQA, with the overall architecture shown in Fig. 1. It consists of three stages: CLIP IQA task guidance, CLIP IQA capability enhancement, and modality-adaptive knowledge distillation.

### A. CLIP IQA Task Guidance

To apply CLIP to IQA tasks, we used a natural language template-guided quality perception method [3] to generate prediction scores related to subjective image quality when adapting CLIP to image-text matching tasks.

Specifically, we adopted the five-point Likert scale widely used in subjective IQA and designed a set of natural language description templates $\mathcal{T}$ related to image quality semantics:

$$\mathcal{T} = \{\text{"a photo of [level] quality"} \mid level \in \mathcal{L}\} \quad (1)$$

where $\mathcal{L} = \{bad, poor, fair, good, perfect\}$ represents the five subjective quality levels of images from low to high. Given an image to be evaluated, we first input it into the CLIP image encoder to obtain its original image features $v_{img} \in \mathbb{R}^{\dim}$. Simultaneously, we input the set of text templates $\mathcal{T}$ corresponding to the five quality levels into the CLIP text encoder, extract the corresponding text features $v_{text}^{(1:5)} \in \mathbb{R}^{\dim}$, and calculate the cosine similarity between the image features and each text template feature:

$$s_i = \cos\langle v_{img}, v_{text}^{(i)} \rangle = \frac{v_{img} \cdot v_{text}^{(i)}}{\|v_{img}\| \|v_{text}^{(i)}\|} \quad (2)$$

The resulting values are divided by a learnable temperature parameter $\tau \in \mathbb{R}^+$ to adjust the smoothness of the similarity distribution, and then undergo Softmax normalization to obtain the probabilities of the image belonging to each quality grade:

$$p_i = \frac{e^{\frac{s_i}{\tau}}}{\sum_{j=1}^{5} e^{\frac{s_j}{\tau}}} \quad (3)$$

This probability distribution $\boldsymbol{p}$ can be regarded as CLIP's perception of the quality level of an image. To map this distribution to a specific quality score, we further introduce a scoring weight vector $\boldsymbol{w} = \{w_i\}_{i=1}^{5} = \{i\}_{i=1}^{5}$ representing the scores corresponding to the five levels. The final prediction quality score $\hat{s}$ can be calculated by weighted summation:

$$\hat{s} = \sum_{i=1}^{5} p_i \cdot w_i \quad (4)$$

However, it should be noted that although CLIP has excellent cross-modal semantic understanding capabilities, when directly applying the above template-guided method, the quality scores predicted by CLIP show low consistency with subjective quality scores across multiple IQA datasets. Therefore, it is necessary to design fine-tuning strategies to enhance its ability to perceive image quality-related features.

### B. CLIP IQA Capability Enhance

Considering that the dataset in the IQA task consists solely of unimodal image-quality score pairs, updating the parameters of the CLIP text encoder during fine-tuning may disrupt its original multimodal semantic alignment. Therefore, we adopted a robust fine-tuning strategy: during fine-tuning, we only updated the parameters of the image encoder, while freezing the parameters of the text encoder and temperature parameters.

The five quality level text templates used in our method are fixed, and the CLIP text encoder remains frozen throughout the process, so the text features corresponding to each text template are also fixed. Therefore, we can preprocess the text features corresponding to each text template, avoiding the need to repeatedly calculate text features during each quality score prediction.

Specifically, we still use the aforementioned guidance mechanism to predict the image quality score. Let the predicted quality score output by CLIP be $\hat{s}$, and its corresponding subjective quality score be $s$. We use mean square error (MSE) as the training objective function:

$$\mathcal{L}_{MSE} = \frac{1}{N}\sum_{i=1}^{N}(\hat{s} - s)^2 \quad (5)$$

*C. Modality-Adaptive Knowledge Distillation*

Although the knowledge learned by CLIP during large-scale pre-training enables it to quickly generalize to IQA tasks, as mentioned earlier, CLIP's limitations restrict its performance. Therefore, we propose a modality-adaptive knowledge distillation strategy, using CLIP as the teacher model and constructing the student model using an image encoder with a small number of parameters that focuses on local feature modeling. This approach enables the teacher model to guide the student model in enhancing its image quality perception capabilities using its rich IQA knowledge. The process primarily consists of three parts: teacher model knowledge extraction, student model design, and knowledge distillation strategy.

**Teacher Model Knowledge Extraction.** The knowledge extracted from our teacher model can be divided into two parts: quality semantic knowledge and quality visual knowledge. Specifically, quality semantic knowledge refers to the textual features of the quality-graded text templates output by the text encoder, which is the model's abstract expression of quality-graded semantics. Quality visual knowledge refers to the image features of the images to be evaluated output by the image encoder, which reflects the model's direct perception of images.

**Student Model Design.** Since the features output by the teacher image encoder are in a specific multimodal semantic space, in order to ensure that the features output by the student image encoder are in the same semantic space, we chose to refer to the original architecture of the teacher model when designing the student model, directly replacing the image encoder with the student image encoder while keeping the rest of the structure unchanged. This also allows the student model to predict image quality scores in the same way as the teacher model.

**Knowledge Distillation Strategy.** The knowledge distillation process aims to guide the training of the student model using the knowledge from the teacher model. For quality semantic knowledge, the student model can directly reuse the text features from the teacher model. For quality visual knowledge, the student model receives both soft label supervision and hard label supervision during the knowledge distillation process, with only the weights of the student encoder being trainable.

The soft label is the feature $v_{img}$ output by the teacher image encoder. For the same input image, soft label supervision maximizes the cosine similarity between the feature $u_{img}$ output by the student model image encoder and $v_{img}$:

$$\mathcal{L}_{soft} = 1 - \frac{1}{N}\sum_{i=1}^{N} cos\langle u_{img}^{(i)}, v_{img}^{(i)}\rangle \quad (6)$$

The hard label is the subjective quality score $s$ in the dataset. Hard label supervision is achieved by minimizing the MSE between the predicted quality score $\hat{s}$ and $s$:

$$\mathcal{L}_{hard} = \frac{1}{N}\sum_{i=1}^{N}(\hat{s} - s)^2 \quad (7)$$

Regarding the contribution of two types of label supervision to the training process, traditional methods [5] mostly use fixed soft and hard label weights, and the overall loss function is defined as:

$$\mathcal{L} = \lambda \cdot \mathcal{L}_{soft} + (1 - \lambda) \cdot \mathcal{L}_{hard} \quad (8)$$

where $\lambda \in (0, 1)$ is a fixed value for the soft label weight that does not change with the number of training iterations. Typically, $\lambda$ is large, ensuring that the teacher model's guidance dominates the training of the student model. To allow the student model to break free from the teacher model's constraints and leverage its own advantages in modeling local features, this chapter proposes a soft-hard label weight scheduling strategy based on cosine annealing:

$$\mathcal{L} = \lambda(t) \cdot \mathcal{L}_{soft} + (1 - \lambda(t)) \cdot \mathcal{L}_{hard} \quad (9)$$

where $\lambda(t)$ denotes the soft label weight value that gradually decreases with training iteration $t$, following a cosine annealing function form:

$$\lambda(t) = \frac{1}{2}(1 + \cos(\frac{t}{T}\pi)) \quad (10)$$

This scheduling strategy achieves a smooth transition from soft label supervision to hard label supervision. In the early stages of training, soft label loss is used to guide the student model to align features, alleviating the initial semantic space differences between the teacher model and the student model while learning the rich IQA knowledge in the teacher model's soft labels. In the later stages of training, hard label loss is used to guide the student model to learn autonomously, allowing the student model to leverage its own architectural advantages.

III. EXPERIMENTS

*A. Experimental Steps*

Our method was evaluated through performance comparison experiments on four mainstream IQA datasets. Among these, CSIQ [7] and LIVE [8] are synthetic distortion IQA datasets, while KonIQ [9] and SPAQ [10] are IQA datasets collected from real-world scenarios. We employed the PLCC and SRCC metrics to quantify the consistency between the model's predicted quality scores and subjective quality scores. We employed the ViT-B/32 version of CLIP and selected several lightweight image encoders proposed in recent years as student models, including Swin Transformer T [11], ResNet 18 [12], MobileViT S [13], and EfficientNet B0 [14], all pre-trained on the ImageNet1K image classification dataset [15]. These are the models with the fewest parameters in their respective series.

TABLE I. THE RESULTS OF PERFORMANCE COMPARISON

|  | CSIQ | | LIVE | | KonIQ | | SPAQ | |
|---|---|---|---|---|---|---|---|---|
|  | PLCC | SROCC | PLCC | SROCC | PLCC | SROCC | PLCC | SROCC |
| BRISQUE[16] | 0.748 | 0.812 | 0.944 | 0.929 | 0.681 | 0.665 | 0.817 | 0.809 |
| NIQE[17] | 0.735 | 0.762 | 0.907 | 0.908 | 0.300 | 0.276 | 0.685 | 0.697 |
| DBCNN[18] | 0.959 | 0.946 | 0.971 | 0.968 | 0.884 | 0.875 | 0.915 | 0.911 |
| TIQA[19] | 0.838 | 0.825 | 0.965 | 0.949 | 0.903 | 0.892 | - | - |
| MetaIQA[20] | 0.908 | 0.899 | 0.959 | 0.960 | 0.856 | 0.887 | - | - |
| HyperIQA(27M)[21] | 0.942 | 0.923 | 0.966 | 0.962 | 0.917 | 0.906 | 0.915 | 0.911 |
| TRes(152M)[22] | 0.942 | 0.922 | 0.968 | 0.969 | 0.928 | 0.915 | - | - |
| MUSIQ(27M)[23] | 0.893 | 0.871 | 0.911 | 0.940 | 0.928 | 0.916 | 0.921 | 0.918 |
| Re-IQA(48M)[24] | 0.960 | 0.947 | 0.971 | 0.970 | 0.923 | 0.914 | 0.925 | 0.918 |
| DEIQT(24M)[25] | 0.963 | 0.946 | 0.982 | 0.980 | 0.934 | 0.921 | 0.923 | 0.919 |
| QFM-IQM(24M)[26] | 0.965 | 0.954 | **0.983** | **0.981** | 0.936 | 0.922 | 0.924 | **0.920** |
| LoDa(118M)[27] | - | - | 0.979 | 0.975 | **0.944** | **0.932** | 0.928 | 0.925 |
| CLIP-IQA(151M)[2] | 0.698 | 0.682 | 0.832 | 0.845 | 0.727 | 0.695 | 0.735 | 0.738 |
| CLIP-IQA+(151M)[2] | 0.887 | 0.894 | 0.912 | 0.917 | 0.909 | 0.895 | 0.866 | 0.864 |
| LIQE(151M)[3] | 0.939 | 0.936 | 0.951 | 0.970 | 0.908 | 0.919 | - | - |
| CLIP→Swin Transformer T(28M) | **0.990** | **0.988** | **0.985** | **0.982** | **0.936** | **0.923** | **0.930** | **0.925** |
| CLIP→ResNet-18(12M) | 0.980 | 0.976 | 0.958 | 0.975 | 0.884 | 0.851 | 0.905 | 0.899 |
| CLIP→MobileViT S(6M) | **0.986** | **0.982** | 0.979 | 0.975 | 0.912 | 0.890 | 0.918 | 0.912 |
| CLIP→EfficientNet B0(5M) | 0.981 | 0.978 | 0.977 | 0.976 | 0.898 | 0.873 | 0.918 | 0.910 |

## B. Implementation details

We selected AdamW as the optimizer. For CLIP and each student model, we set the initial learning rate to 5e-6 and 1e-4, respectively. For all models, we chose to decrease the learning rate according to a cosine curve, which decreased to 0.1 times the initial learning rate in the final stage. All models are trained for 100 epochs, with a batch size of 64. The input images are fixed to a size of 224x224, and the student models use the same image preprocessing function as CLIP to align the data distribution. Each dataset is randomly split into 80% for training and 20% for testing. We conduct multiple experiments and take the median as the final performance result of the model. The experiments are conducted on a single NVIDIA RTX 4090D.

## C. Main Results

To comprehensively evaluate the performance of our method, we applied our proposed method to the student image encoders we selected and compared it with other existing methods. The compared methods include those based on handcrafted features and those based on deep learning, with CLIP-based methods identified separately, as shown in Table I. The number of parameters for each model is indicated after the name, and the top two for each dataset are highlighted in bold. The results demonstrate that our proposed method achieves competitive performance, performing exceptionally well in both synthetic and real-world distortion evaluations.

## D. Ablation Studies

To assess the importance of each component in the proposed method, we propose several variant models: (1) directly concatenating a regression head to the features output by the student image encoder to output a quality score; (2) using only hard labels in knowledge distillation; (3) using only soft labels in knowledge distillation; (4) using our proposed complete knowledge distillation strategy. Their performance is shown in Table II, where A to D represent the four student image encoders, in the same order as in Table I. The results show that for each student model, (4) outperforms all other variants, and the weaker the performance of (1), the greater the performance improvement of (4). Overall, the (4) variant of all student models achieves an average improvement of 0.016 in PLCC, respectively, compared to (1) across all datasets. This indicates that each component of our proposed method plays a crucial role.

TABLE II. THE PLCC RESULTS OF ABLATION STUDIES

|  | CSIQ | LIVE | KonIQ | SPAQ |
|---|---|---|---|---|
| A | 0.987 | 0.981 | 0.926 | 0.927 |
| A+hard label | 0.841 | 0.789 | 0.831 | 0.858 |
| A+soft label | 0.977 | 0.949 | 0.926 | 0.927 |
| A+our strategy | **0.990** | **0.985** | **0.936** | **0.930** |
| B | 0.968 | 0.955 | 0.843 | 0.893 |
| B+hard label | 0.832 | 0.782 | 0.775 | 0.836 |
| B+soft label | 0.967 | 0.924 | 0.881 | 0.903 |
| B+our strategy | **0.980** | **0.958** | **0.884** | **0.905** |
| C | 0.975 | 0.962 | 0.897 | 0.909 |
| C+hard label | 0.817 | 0.766 | 0.810 | 0.843 |
| C+soft label | 0.971 | 0.940 | 0.904 | 0.914 |
| C+our strategy | **0.986** | **0.979** | **0.912** | **0.918** |
| D | 0.959 | 0.947 | 0.860 | 0.898 |
| D+hard label | 0.826 | 0.772 | 0.797 | 0.845 |
| D+soft label | 0.968 | 0.945 | 0.887 | 0.910 |
| D+our strategy | **0.981** | **0.977** | **0.898** | **0.918** |

## IV. CONCLUSION

This paper proposes a knowledge distillation method based on a vision-language model for image quality assessment. By leveraging the guidance and capability enhancement of the CLIP IQA task and employing an adaptive distillation strategy, it effectively addresses the issues of excessive parameter counts and insufficient ability to identify local distortion features in CLIP for IQA. Each component of the adaptive distillation method has been validated through ablation experiments. Experimental results demonstrate that the method achieves excellent evaluation metrics on mainstream IQA datasets while significantly reducing the number of model parameters, showcasing strong practical application potential.